\documentclass[a4paper,fleqn]{cas-sc}
\usepackage[numbers,sort&compress]{natbib}
\usepackage{amsmath,amssymb,bm,mathtools}
\usepackage{graphicx}
\usepackage{booktabs}
\usepackage{longtable}
\usepackage{multirow}
\usepackage{array}
\usepackage{float}
\usepackage{makecell}
\usepackage{caption}
\usepackage[section]{placeins}
\usepackage{hyperref}
\graphicspath{{./}}
\newcommand{\Eq}[1]{Eq.(\ref{#1})}
\newcommand{\Fig}[1]{Fig.(\ref{#1})}
\newcommand{\Tab}[1]{Table \ref{#1}}
\newcommand{\SMSec}[1]{SM Section~#1}
\newcommand{\bx}{\bm{x}}
\newcommand{\bn}{\bm{n}}
\newcommand{\bU}{\bm{U}}
\newcommand{\bG}{\bm{G}}

\newcommand{\dd}{\mathrm{d}}

\begin{document}
\let\WriteBookmarks\relax

\shorttitle{RA-PINN for Steady Electrothermal Multiphysics}
\shortauthors{Y.Zhou et~al.}

\title[mode=title]{Residual Attention Physics-Informed Neural Networks for Robust Multiphysics Simulation of Steady-State Electrothermal Energy Systems}

\author[1]{Yuqing Zhou}[orcid=0009-0005-5038-6604]
\credit{Calculation, data analyzing and manuscript writing}
\fnref{co-first}\author[1]{Ze Tao}[orcid=0009-0004-0202-3641]
\credit{Calculation, data analyzing, manuscript writing, review and editing}
\fnref{co-first}
\affiliation[1]{organization={Nanophotonics and Biophotonics Key Laboratory of Jilin Province, School of Physics, Changchun University of Science and Technology},
                city={Changchun},
%               citysep={}, % Uncomment if no comma needed between city and postcode
                postcode={130022},
                country={P.R. China}}
\author[1]{Fujun Liu}[orcid=0000-0002-8573-450X]
\corref{cor}
\credit{Review and Editing}
%\cormark[2]
\ead{fjliu@cust.edu.cn}
% Corresponding author text
\fntext[co-first]{The authors contribute equally to this work.}
\cortext[cor]{Corresponding author}

\begin{abstract}
Efficient thermal management and precise field prediction are critical for the design of advanced energy systems, including electrohydrodynamic transport, microfluidic energy harvesters, and electrically driven thermal regulators. However, the steady-state simulation of these electrothermal coupled multiphysics systems remains challenging for physics-informed neural computation due to strong nonlinear field coupling, temperature-dependent coefficient variability, and complex interface dynamics. This study proposes a Residual Attention Physics-Informed Neural Network (RA-PINN) framework for the unified solution of coupled velocity, pressure, electric-potential, and temperature fields. By integrating a unified five-field operator formulation with residual-connected feature propagation and attention-guided channel modulation, the proposed architecture effectively captures localized coupling structures and steep gradients. We evaluate RA-PINN across four representative energy-relevant benchmarks: constant-coefficient coupling, indirect pressure-gauge constraints, temperature-dependent transport, and oblique-interface consistency. Comparative analysis against Pure-MLP, LSTM-PINN, and pLSTM-PINN demonstrates that RA-PINN achieves superior accuracy, yielding the lowest MSE, RMSE, and relative $L_2$ errors across all scenarios. Notably, RA-PINN maintains high structural fidelity in interface-dominated and variable-coefficient settings where conventional PINN backbones often fail. These results establish RA-PINN as a robust and accurate computational framework for the high-fidelity modeling and optimization of complex electrothermal multiphysics in sustainable energy applications.
\end{abstract}

\begin{highlights}
\item A unified five-field formulation resolves coupled electrothermal energy transport.
\item Residual-attention PINN captures localized gradients in complex energy domains.
\item Adaptive sampling enforces constraints on interfaces and variable coefficients.
\item RA-PINN yields superior accuracy across all benchmarks compared to reference PINNs.
\item Enhanced robustness is achieved for thermal management and microenergy systems.
\end{highlights}


\maketitle

\section{Introduction}
Partial differential equations (PDEs) provide the fundamental mathematical framework for governing electric transport, viscous motion, pressure redistribution, and heat transfer across a diverse array of coupled scientific and engineering systems. In many practical applications, these mechanisms interact within the same spatial domain, causing the solution to evolve beyond a simple smooth scalar field. Consequently, velocity, pressure, electric potential, and temperature must satisfy a tightly coupled constraint system where each field feeds back into the others through transport, forcing, and constitutive effects. This inherent complexity gives steady electrothermal coupled multiphysics problems significant scientific value and direct engineering relevance, as accurate field prediction is essential for device design, parameter tuning, and stability analysis in electrohydrodynamic transport, electrothermal regulation, and electrically driven microenergy systems \citep{raissi2019pinn,zhao2025heatreview}.
Recent studies in \emph{Applied Energy} further confirm the practical importance of thermally coupled prediction and control in energy applications, including building thermal modeling, liquid-cooled battery systems, thermoelectric battery thermal management, and physics-informed battery health estimation \citep{dinatale2022pcnn,wu2024liquidcooling,luo2024btms,yang2025batterypinn}.

To address these challenges, researchers have historically utilized mesh-based solvers and, more recently, a broad class of physics-informed neural network (PINN) variants targeting stability, weighting, and collocation design. PINNs embed governing equations, boundary conditions, and auxiliary constraints directly into the training objective, effectively transforming PDE solving into a constrained function approximation problem \citep{raissi2019pinn}. Subsequent studies have sought to overcome the persistent weaknesses of vanilla PINNs through optimization analysis, self-adaptive weighting, and residual-based adaptivity \citep{urban2025optimization,chen2025balanced,toscano2026variational}. Despite these advances, strongly coupled steady multiphysics problems still present significant difficulties. Physical fields often exhibit vastly different magnitudes and gradient scales, which can cause the optimization process to privilege one field while suppressing another. Furthermore, variable coefficients and indirect constraints can stiffen the loss landscape, while interface-dominated solutions force the network to balance broad, smooth transport against narrow, local transitions.

Residual-attention modeling offers a credible structural remedy for these optimization imbalances. Recent PINN studies have successfully employed residual-based attention, loss-attentional weighting, and residual-connected architectures to emphasize hard-to-fit regions and stabilize feature transport during training \citep{anagnostopoulos2024rba,song2024lapinn,tian2024residual,ramirez2025transformer}. By combining these structural innovations with balanced weighting and residual-based adaptivity, optimization can be significantly improved in historically stubborn regions \citep{chen2025balanced,toscano2026variational,urban2025optimization}. Within this framework, residual pathways preserve deep-feature transport while attention gates amplify informative channels that encode steep transitions, localized coupling structures, and interface-sensitive signatures. Rather than replacing physical constraints, this combination enriches the representation space that the physics-informed optimizer explores, thereby enhancing the capacity of the model to resolve complex multiphysics interactions.

The broader literature reflects a rapidly expanding landscape of physics-informed learning designed to tackle such complexities. Researchers have developed multiresolution PDE-preserving learning, physics-informed neural operators, and hybrid analytical-neural solvers for nonlinear heat conduction \citep{liu2024multires,li2024pino,jiao2025oneshot,zhao2025dno,tao2025analytical}. Specialized studies have further targeted long-horizon thermal prediction in manufacturing, steady electrohydrodynamic LSTM-PINN modeling, interval-constrained interface dynamics, and digital-twin-oriented workflows \citep{tian2025manufacturing,moon2025thermoelectric,xing2025modeling,rezaei2025digitalizing,yang2024digitaltwin}. Advances in architecture and discretization now include finite-element-interpolated networks, multilevel domain decomposition, and temporally continuous sequential PINNs \citep{badia2024feinn,dolean2024multilevel,roy2024temporal,he2024multilevel,shukla2024fairkan}. Concurrent efforts have also addressed high-dimensional learning difficulties, automatic structure discovery, and learnable activation functions \citep{hu2024curse,kiyani2025optimizer,liu2025structure,farea2025laf,sharma2026graph}. Furthermore, problem-specific extensions now cover physics-informed transformers, unified physics-only training, and coordinate-transformed or geometry-aware extensions \citep{sobral2025transformers,tao2025lnn,zheng2026igpinn,zhang2024pipnn,liu2025icpinn}. Additional studies have explored boundary-layer transformations, hyperelastic response, compatible feature constructions, and residual-attention mechanisms \citep{zhang2025boundarylayer,moon2025hyperelastic,badia2025compatible,tian2024residual}.

Motivated by these developments, we apply the Residual Attention Physics-Informed Neural Network (RA-PINN) to four steady electrothermal benchmark cases, providing a rigorous comparison with Pure-MLP, LSTM-PINN, and pLSTM-PINN solvers. The governing system is formulated in a unified operator framework, allowing for a consistent analysis of constant-coefficient, pressure-gauge, temperature-dependent, and oblique-interface configurations. Our evaluation summarizes field-wise and case-average metrics including MSE, RMSE, MAE, and relative $L_2$ error, alongside training time requirements. The results demonstrate that RA-PINN effectively addresses the limitations of conventional backbones in steady coupled electrothermal simulations, particularly when the solution is dominated by local structures, indirect constraints, or coefficient heterogeneity \citep{anagnostopoulos2024rba,song2024lapinn,toscano2026variational}. Detailed technical support, including notation, solver decomposition, unified residual operators, and adaptive sampling protocols, is provided in the accompanying supplementary material (SM Sections I–VIII).

\section{RA-PINN Architecture}
\begin{figure}[t]
    \centering
    \includegraphics[width=0.98\textwidth]{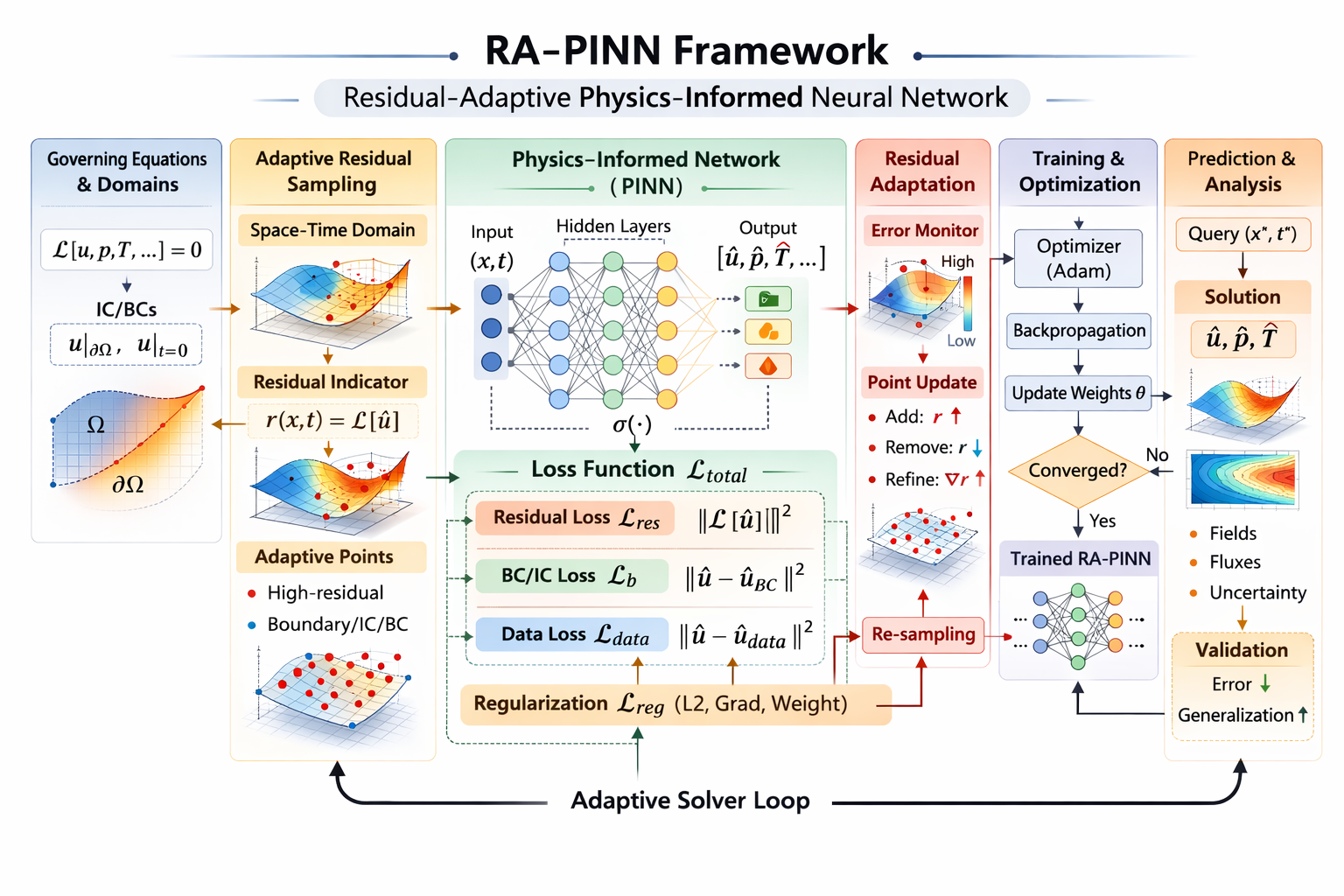}
    \caption{Workflow of the RA-PINN solver used in this study. The schematic organizes the method into six coupled parts: governing equations and computational domain, adaptive residual sampling, residual-attention PINN mapping, residual adaptation, gradient-based optimization, and prediction/validation. The uploaded schematic uses generic space-time notation; in the present steady-state problem the network input reduces to the spatial coordinate $(x,y)$ and the outputs reduce to the five coupled fields $\hat u$, $\hat v$, $\hat p$, $\hat \phi$, and $\hat T$.}
    \label{fig:ra_pinn_arch}
\end{figure}
\FloatBarrier

\Fig{fig:ra_pinn_arch} gives a complete view of how RA-PINN enters the present steady electrothermal study. The present solver combines two literature-supported ingredients: a residual-connected deep backbone and an attention-guided residual-focusing strategy motivated by recent PINN studies on residual connections and residual-based attention \citep{tian2024residual,anagnostopoulos2024rba,ramirez2025transformer}. The leftmost block starts from the governing PDE system, the computational domain $\Omega$, and the boundary constraints. These ingredients define the physical admissible set of the solution. The next block constructs an adaptive residual-sampling mechanism. Candidate interior and boundary points enter the residual indicator, and the solver identifies the locations where the PDE residual remains high. This step matters because the coupled pressure, velocity, electric-potential, and temperature fields do not distribute their errors uniformly in space. Sharp transitions, coefficient-sensitive regions, and interface neighborhoods usually concentrate the dominant approximation error, so the collocation set should move toward these locations instead of remaining fixed during the whole optimization process.

In the present steady problem, the RA-PINN backbone maps the spatial coordinate $\bx=(x,y)^\top\in\Omega$ to the five-field prediction vector
\begin{equation}
\widehat{\bU}(\bx;\theta)=\mathcal{F}_{\theta}(\bx)=\left[\hat u(\bx),\hat v(\bx),\hat p(\bx),\hat \phi(\bx),\hat T(\bx)\right]^\top,
\label{eq:ra_map}
\end{equation}
where $\theta$ collects all trainable parameters. The schematic labels the input as $(x,t)$ because it describes a generic PINN template, but the present paper solves steady problems only, so the temporal coordinate does not appear in the actual network input. Inside the backbone, the residual pathway transports low-level and mid-level features directly across depth, while the attention pathway performs channel-wise modulation and strengthens the features that carry physically informative structures.

Let $\bm{z}^{(\ell)}$ denote the latent feature at the $\ell$th block. The trunk branch and the attention branch generate
\begin{equation}
\bm{t}^{(\ell)}=\mathcal{T}^{(\ell)}\!\left(\bm{z}^{(\ell)}\right),
\qquad
\bm{m}^{(\ell)}=\sigma\!\left(\mathcal{M}^{(\ell)}\!\left(\bm{z}^{(\ell)}\right)\right),
\label{eq:tm}
\end{equation}
and the block output satisfies
\begin{equation}
\bm{z}^{(\ell+1)}=\bm{z}^{(\ell)}+\left(\bm{1}+\bm{m}^{(\ell)}\right)\odot\bm{t}^{(\ell)}.
\label{eq:resattn}
\end{equation}
\Eq{eq:resattn} shows the core mechanism of RA-PINN. The residual connection keeps gradient transmission stable and preserves global background information, while the attention gate reweights channels before fusion and thereby improves the network response to localized structures. For steady electrothermal coupling, this property directly helps the solver resolve narrow pressure layers, temperature-dependent distortions, and oblique interface signatures without sacrificing the smooth large-scale field profile. \SMSec{II} distinguishes the block-level modulation from residual-driven collocation focusing, and recent studies on residual-based attention and loss-attentional weighting provide the broader PINN perspective for this design choice \citep{anagnostopoulos2024rba,song2024lapinn,ramirez2025transformer}.

After the network produces $\widehat{\bU}$, automatic differentiation constructs the residual vector associated with the governing operator. We write the pointwise residual indicator as
\begin{equation}
r(\bx;\theta)=\left\|\mathcal{N}\!\left(\widehat{\bU}(\bx;\theta)\right)\right\|_2,
\label{eq:res_indicator}
\end{equation}
where $\mathcal{N}(\cdot)$ denotes the coupled PDE operator introduced in Section~3. The adaptive residual-sampling block and the residual-adaptation block in \Fig{fig:ra_pinn_arch} use $r(\bx;\theta)$ to update the collocation distribution. Points with persistently large residuals receive more attention in the next optimization round, whereas low-residual points can leave the active set or enter it with lower density. This strategy forms the adaptive solver loop shown at the bottom of the schematic and lets the network spend more capacity on the dynamically hardest regions of the domain. \SMSec{II} states the solver decomposition, \SMSec{V} writes the adaptive collocation rule in algorithmic form, and \SMSec{VIII} records the object-level implementation correspondence used throughout the experiments.

The loss-function panel in \Fig{fig:ra_pinn_arch} also matches the present formulation. We minimize a composite objective of the form
\begin{equation}
\mathcal{L}_{\mathrm{total}}=\lambda_{\mathrm{res}}\mathcal{L}_{\mathrm{res}}+\lambda_{\mathrm{b}}\mathcal{L}_{\mathrm{b}}+\lambda_{\mathrm{data}}\mathcal{L}_{\mathrm{data}}+\lambda_{\mathrm{reg}}\mathcal{L}_{\mathrm{reg}}+\lambda_{\mathrm{gauge}}\mathcal{L}_{\mathrm{gauge}}+\lambda_{\Gamma}\mathcal{L}_{\Gamma},
\label{eq:loss_total}
\end{equation}
where $\mathcal{L}_{\mathrm{res}}$ penalizes the interior PDE residuals, $\mathcal{L}_{\mathrm{b}}$ enforces boundary conditions, $\mathcal{L}_{\mathrm{data}}$ denotes the supervised term when reference samples participate in training, $\mathcal{L}_{\mathrm{reg}}$ denotes regularization, $\mathcal{L}_{\mathrm{gauge}}$ activates the pressure-gauge constraint when needed, and $\mathcal{L}_{\Gamma}$ enforces interface continuity and jump consistency in the interface case. Gradient-based optimization then updates the parameter vector according to
\begin{equation}
\theta^{k+1}=\theta^{k}-\eta_k\nabla_{\theta}\mathcal{L}_{\mathrm{total}}\!\left(\theta^{k}\right),
\label{eq:adam_update}
\end{equation}
which corresponds to the training-and-optimization block in \Fig{fig:ra_pinn_arch}. After convergence, the trained RA-PINN outputs the complete field solution together with the post-processed error analysis. \SMSec{IV} records the detailed loss decomposition, the pressure-gauge activation rule, and the interface-continuity and flux channels, \SMSec{V} gives the interface sampling construction, and \SMSec{VIII} links these mathematical objects to the executable solver blocks. Therefore, the uploaded architecture figure does not merely illustrate a generic deep network. It precisely matches the present workflow in which physical constraints, residual-attention representation learning, and adaptive point redistribution cooperate inside one steady electrothermal PINN solver.

\section{Unified PDE Formulation}
We write the steady electrothermal coupled system in one operator form over the unit square domain $\Omega=[0,1]\times[0,1]$. The structured-grid notation that accompanies the benchmark data is summarized in \SMSec{I}. The unknown vector reads
\begin{equation}
\bU(x,y)=\left[u(x,y),v(x,y),p(x,y),\phi(x,y),T(x,y)\right]^\top.
\label{eq:U}
\end{equation}
We denote the two-dimensional gradient by $\nabla=(\partial_x,\partial_y)^\top$ and the Laplacian by $\Delta=\partial_{xx}+\partial_{yy}$.

The steady incompressibility constraint reads
\begin{equation}
R_c(\bU)=\nabla\cdot\bm{u}=\partial_x u+\partial_y v=0,
\label{eq:cont}
\end{equation}
where $\bm{u}=(u,v)^\top$. We write the momentum equations as
\begin{equation}
\bm{R}_{\mathrm{mom}}(\bU)=\rho\,(\bm{u}\cdot\nabla)\bm{u}+\nabla p-\nabla\cdot\!\left(\nu\nabla\bm{u}\right)-\bm{f}_{\mathrm{e}}(\phi,T)=\bm{0},
\label{eq:mom}
\end{equation}
where $\rho$ denotes density, $\nu$ denotes the viscosity-related transport coefficient, and $\bm{f}_{\mathrm{e}}$ denotes the electrothermal body-force contribution.

The electric-potential equation reads
\begin{equation}
R_{\phi}(\bU)=-\nabla\cdot\!\left(\sigma\nabla\phi\right)-s_{\phi}(x,y)=0,
\label{eq:phi_eq}
\end{equation}
where $\sigma$ denotes electric conductivity and $s_{\phi}$ denotes the source term selected by the benchmark.

The temperature equation reads
\begin{equation}
R_T(\bU)=-\nabla\cdot\!\left(\alpha\nabla T\right)+\bm{u}\cdot\nabla T-s_T(x,y)=0,
\label{eq:T_eq}
\end{equation}
where $\alpha$ denotes thermal diffusivity and $s_T$ denotes the thermal source term.

We collect the full interior residual in one vector operator,
\begin{equation}
\mathcal{N}(\bU)=\left[R_c,\,R_u,\,R_v,\,R_{\phi},\,R_T\right]^\top=\bm{0},
\label{eq:operator}
\end{equation}
where $R_u$ and $R_v$ denote the two scalar components of \Eq{eq:mom}. This notation lets us keep one PDE template across all four cases.

The cases differ through the constraint operator and the coefficient law. In the constant-coefficient benchmarks we keep $\nu$, $\alpha$, $\sigma$, and the coupling coefficients constant. In the pressure-gauge benchmark we replace direct pressure anchoring with
\begin{equation}
\int_{\Omega} p\,\dd\Omega=0.
\label{eq:gauge}
\end{equation}
In the temperature-dependent benchmark we activate variable transport coefficients,
\begin{equation}
\nu(T)=\nu_0\left(1+\beta_{\nu}T\right),
\qquad
\alpha(T)=\alpha_0\left(1+\beta_{\alpha}T\right).
\label{eq:tempdep}
\end{equation}
In the interface benchmark we split the domain by the oblique interface
\begin{equation}
\Gamma=\left\{(x,y)\in\Omega\,\middle|\,g(x,y)=x+0.4y-0.7=0\right\},
\label{eq:interface}
\end{equation}
with subdomains $\Omega^- = \{g<0\}$ and $\Omega^+ = \{g>0\}$. We then enforce field continuity and flux consistency across $\Gamma$,
\begin{equation}
[u]=[v]=[p]=[\phi]=[T]=0,
\qquad
[\nu\nabla u\cdot\bn],\;[\nu\nabla v\cdot\bn],\;[\alpha\nabla T\cdot\bn],\;[\sigma\nabla\phi\cdot\bn]\ \text{follow the prescribed jumps}.
\label{eq:jump}
\end{equation}
We next write one unified residual construction for all benchmark cases and link it directly to \Eq{eq:loss_total}. Let $\widehat{\bU}(x,y;\theta)$ denote the network prediction. We define the interior residual vector by
\begin{equation}
\mathcal{R}_{\Omega}(x,y;\theta)=\mathcal{N}(\widehat{\bU}(x,y;\theta)),
\qquad (x,y)\in \Omega,
\label{eq:res_omega}
\end{equation}
and we define the boundary residual by
\begin{equation}
\mathcal{R}_{\partial\Omega}(x,y;\theta)=\mathcal{B}(\widehat{\bU}(x,y;\theta))-\bG(x,y),
\qquad (x,y)\in \partial\Omega,
\label{eq:res_boundary}
\end{equation}
where $\mathcal{B}$ collects the prescribed boundary operators and $\bG$ collects the corresponding boundary data. In the pressure-gauge case, we add the global residual
\begin{equation}
\mathcal{R}_{\mathrm{g}}(\theta)=\int_{\Omega} \hat p(x,y;\theta)\,\dd\Omega.
\label{eq:res_gauge}
\end{equation}
In the interface case, we add the transmission residual
\begin{equation}
\mathcal{R}_{\Gamma}(x,y;\theta)=
\left[
[u],[v],[p],[\phi],[T],[\nu\nabla u\cdot\bn],[\nu\nabla v\cdot\bn],[\alpha\nabla T\cdot\bn],[\sigma\nabla\phi\cdot\bn]
\right]^\top,
\qquad (x,y)\in \Gamma.
\label{eq:res_interface}
\end{equation}
We then write the case-wise residual objective in the unified form
\begin{equation}
\mathcal{L}_{\mathrm{case}}=\lambda_{\mathrm{res}}\,\|\mathcal{R}_{\Omega}\|_2^2+\lambda_{\mathrm{b}}\,\|\mathcal{R}_{\partial\Omega}\|_2^2+\lambda_{\mathrm{gauge}}\,|\mathcal{R}_{\mathrm{g}}|^2+\lambda_{\Gamma}\,\|\mathcal{R}_{\Gamma}\|_2^2,
\label{eq:case_residual}
\end{equation}
where the coefficients $\lambda_{\mathrm{res}}$, $\lambda_{\mathrm{b}}$, $\lambda_{\mathrm{gauge}}$, and $\lambda_{\Gamma}$ use the same notation as the corresponding physics terms in \Eq{eq:loss_total}. This choice makes the residual construction and the training objective fully consistent across all benchmark cases. More specifically,
\begin{equation}
\lambda_{\mathrm{res}}\mathcal{L}_{\mathrm{res}}=\lambda_{\mathrm{res}}\,\|\mathcal{R}_{\Omega}\|_2^2,
\quad
\lambda_{\mathrm{b}}\mathcal{L}_{\mathrm{b}}=\lambda_{\mathrm{b}}\,\|\mathcal{R}_{\partial\Omega}\|_2^2,
\quad
\lambda_{\mathrm{gauge}}\mathcal{L}_{\mathrm{gauge}}=\lambda_{\mathrm{gauge}}\,|\mathcal{R}_{\mathrm{g}}|^2,
\quad
\lambda_{\Gamma}\mathcal{L}_{\Gamma}=\lambda_{\Gamma}\,\|\mathcal{R}_{\Gamma}\|_2^2,
\label{eq:loss_link}
\end{equation}
so the physics-dependent part of \Eq{eq:loss_total} reduces to
\begin{equation}
\mathcal{L}_{\mathrm{total}}=\mathcal{L}_{\mathrm{case}}+\lambda_{\mathrm{data}}\mathcal{L}_{\mathrm{data}}+\lambda_{\mathrm{reg}}\mathcal{L}_{\mathrm{reg}}.
\label{eq:loss_total_case}
\end{equation}
This notation keeps one residual template across all four cases while still allowing pressure-gauge and interface constraints to enter only when the corresponding physics requires them. \SMSec{I}, \SMSec{III}, and \SMSec{IV} expand this concise operator-level presentation into the common benchmark notation, the sampled residual channels, and the case-activation rules.

For all four uploaded cases, the benchmark builder prescribes the source terms, coefficients, and boundary or interface data together with reference field data on the evaluation grid. We therefore solve the same PDE system and then compare the learned fields with the provided benchmark field maps and the logged quantitative metrics. The extended RA-PINN solver description, the adaptive residual-focusing rule, the metric definitions, and the executable inventory used in this study are documented in \SMSec{II}--\SMSec{VIII}.

\begingroup
\footnotesize
\setlength{\LTleft}{0pt}
\setlength{\LTright}{0pt}
\begin{longtable}{ccccccc}
\caption{Fieldwise and average error table for all four benchmark cases. Each model reports the five physical fields $u$, $v$, $p$, $T$, and $\phi$, together with an average row over the five fields. The minimum value for each case, field, and metric appears in bold.}\\
\label{tab:error_all}\\
\toprule
Case & Model & Field & MSE & RMSE & MAE & Relative $L_2$ \\
\midrule
\endfirsthead
\toprule
Case & Model & Field & MSE & RMSE & MAE & Relative $L_2$ \\
\midrule
\endhead
\midrule
\multicolumn{7}{r}{Continued on next page}\\
\midrule
\endfoot
\bottomrule
\endlastfoot
Case 1 & RA-PINN & u & $\mathbf{8.283\times10^{-7}}$ & $\mathbf{9.101\times10^{-4}}$ & $\mathbf{7.095\times10^{-4}}$ & $\mathbf{2.661\times10^{-3}}$ \\
 &  & v & $\mathbf{7.918\times10^{-7}}$ & $\mathbf{8.898\times10^{-4}}$ & $\mathbf{6.665\times10^{-4}}$ & $\mathbf{3.256\times10^{-3}}$ \\
 &  & p & $\mathbf{7.026\times10^{-7}}$ & $\mathbf{8.382\times10^{-4}}$ & $\mathbf{6.588\times10^{-4}}$ & $\mathbf{4.756\times10^{-3}}$ \\
 &  & T & $\mathbf{1.374\times10^{-6}}$ & $\mathbf{1.172\times10^{-3}}$ & $\mathbf{8.733\times10^{-4}}$ & $\mathbf{2.221\times10^{-3}}$ \\
 &  & phi & $\mathbf{8.446\times10^{-7}}$ & $\mathbf{9.190\times10^{-4}}$ & $\mathbf{5.973\times10^{-4}}$ & $\mathbf{3.279\times10^{-3}}$ \\
 &  & Avg. & $\mathbf{9.083\times10^{-7}}$ & $\mathbf{9.459\times10^{-4}}$ & $\mathbf{7.011\times10^{-4}}$ & $\mathbf{3.235\times10^{-3}}$ \\
\midrule
 & LSTM-PINN & u & $2.265\times10^{-6}$ & $1.505\times10^{-3}$ & $1.144\times10^{-3}$ & $4.400\times10^{-3}$ \\
 &  & v & $1.906\times10^{-6}$ & $1.380\times10^{-3}$ & $1.049\times10^{-3}$ & $5.052\times10^{-3}$ \\
 &  & p & $2.279\times10^{-6}$ & $1.510\times10^{-3}$ & $1.127\times10^{-3}$ & $8.566\times10^{-3}$ \\
 &  & T & $5.023\times10^{-6}$ & $2.241\times10^{-3}$ & $1.684\times10^{-3}$ & $4.246\times10^{-3}$ \\
 &  & phi & $3.030\times10^{-6}$ & $1.741\times10^{-3}$ & $1.143\times10^{-3}$ & $6.210\times10^{-3}$ \\
 &  & Avg. & $2.901\times10^{-6}$ & $1.675\times10^{-3}$ & $1.230\times10^{-3}$ & $5.695\times10^{-3}$ \\
\midrule
 & pLSTM-PINN & u & $1.491\times10^{-5}$ & $3.861\times10^{-3}$ & $3.048\times10^{-3}$ & $1.129\times10^{-2}$ \\
 &  & v & $1.441\times10^{-5}$ & $3.796\times10^{-3}$ & $2.952\times10^{-3}$ & $1.389\times10^{-2}$ \\
 &  & p & $1.056\times10^{-5}$ & $3.249\times10^{-3}$ & $2.358\times10^{-3}$ & $1.844\times10^{-2}$ \\
 &  & T & $8.676\times10^{-6}$ & $2.946\times10^{-3}$ & $2.296\times10^{-3}$ & $5.580\times10^{-3}$ \\
 &  & phi & $9.636\times10^{-6}$ & $3.104\times10^{-3}$ & $2.237\times10^{-3}$ & $1.108\times10^{-2}$ \\
 &  & Avg. & $1.164\times10^{-5}$ & $3.391\times10^{-3}$ & $2.578\times10^{-3}$ & $1.205\times10^{-2}$ \\
\midrule
 & Pure-MLP & u & $1.658\times10^{-4}$ & $1.288\times10^{-2}$ & $9.746\times10^{-3}$ & $3.765\times10^{-2}$ \\
 &  & v & $2.267\times10^{-4}$ & $1.506\times10^{-2}$ & $1.080\times10^{-2}$ & $5.510\times10^{-2}$ \\
 &  & p & $1.990\times10^{-4}$ & $1.411\times10^{-2}$ & $1.106\times10^{-2}$ & $8.004\times10^{-2}$ \\
 &  & T & $3.649\times10^{-4}$ & $1.910\times10^{-2}$ & $1.528\times10^{-2}$ & $3.619\times10^{-2}$ \\
 &  & phi & $1.681\times10^{-4}$ & $1.297\times10^{-2}$ & $1.009\times10^{-2}$ & $4.626\times10^{-2}$ \\
 &  & Avg. & $2.249\times10^{-4}$ & $1.482\times10^{-2}$ & $1.139\times10^{-2}$ & $5.105\times10^{-2}$ \\
\midrule
Case 2 & RA-PINN & u & $\mathbf{7.835\times10^{-7}}$ & $\mathbf{8.852\times10^{-4}}$ & $\mathbf{6.938\times10^{-4}}$ & $\mathbf{2.588\times10^{-3}}$ \\
 &  & v & $\mathbf{6.340\times10^{-7}}$ & $\mathbf{7.963\times10^{-4}}$ & $\mathbf{6.284\times10^{-4}}$ & $\mathbf{2.914\times10^{-3}}$ \\
 &  & p & $\mathbf{6.309\times10^{-6}}$ & $\mathbf{2.512\times10^{-3}}$ & $\mathbf{2.355\times10^{-3}}$ & $\mathbf{2.829\times10^{-2}}$ \\
 &  & T & $\mathbf{2.333\times10^{-6}}$ & $\mathbf{1.527\times10^{-3}}$ & $\mathbf{1.121\times10^{-3}}$ & $\mathbf{2.894\times10^{-3}}$ \\
 &  & phi & $\mathbf{2.052\times10^{-7}}$ & $\mathbf{4.530\times10^{-4}}$ & $\mathbf{3.308\times10^{-4}}$ & $\mathbf{1.616\times10^{-3}}$ \\
 &  & Avg. & $\mathbf{2.053\times10^{-6}}$ & $\mathbf{1.235\times10^{-3}}$ & $\mathbf{1.026\times10^{-3}}$ & $\mathbf{7.660\times10^{-3}}$ \\
\midrule
 & LSTM-PINN & u & $1.778\times10^{-6}$ & $1.333\times10^{-3}$ & $1.023\times10^{-3}$ & $3.899\times10^{-3}$ \\
 &  & v & $2.131\times10^{-6}$ & $1.460\times10^{-3}$ & $1.111\times10^{-3}$ & $5.342\times10^{-3}$ \\
 &  & p & $8.451\times10^{-6}$ & $2.907\times10^{-3}$ & $2.557\times10^{-3}$ & $3.274\times10^{-2}$ \\
 &  & T & $4.804\times10^{-6}$ & $2.192\times10^{-3}$ & $1.681\times10^{-3}$ & $4.152\times10^{-3}$ \\
 &  & phi & $2.615\times10^{-6}$ & $1.617\times10^{-3}$ & $1.091\times10^{-3}$ & $5.769\times10^{-3}$ \\
 &  & Avg. & $3.956\times10^{-6}$ & $1.902\times10^{-3}$ & $1.493\times10^{-3}$ & $1.038\times10^{-2}$ \\
\midrule
 & pLSTM-PINN & u & $4.019\times10^{-5}$ & $6.339\times10^{-3}$ & $5.053\times10^{-3}$ & $1.821\times10^{-2}$ \\
 &  & v & $6.604\times10^{-5}$ & $8.126\times10^{-3}$ & $6.473\times10^{-3}$ & $2.925\times10^{-2}$ \\
 &  & p & $1.464\times10^{-4}$ & $1.210\times10^{-2}$ & $1.032\times10^{-2}$ & $7.171\times10^{-2}$ \\
 &  & T & $4.381\times10^{-5}$ & $6.619\times10^{-3}$ & $5.159\times10^{-3}$ & $1.203\times10^{-2}$ \\
 &  & phi & $1.811\times10^{-4}$ & $1.346\times10^{-2}$ & $1.044\times10^{-2}$ & $4.918\times10^{-2}$ \\
 &  & Avg. & $9.551\times10^{-5}$ & $9.329\times10^{-3}$ & $7.488\times10^{-3}$ & $3.608\times10^{-2}$ \\
\midrule
 & Pure-MLP & u & $1.040\times10^{-4}$ & $1.020\times10^{-2}$ & $8.364\times10^{-3}$ & $2.981\times10^{-2}$ \\
 &  & v & $1.046\times10^{-4}$ & $1.023\times10^{-2}$ & $8.039\times10^{-3}$ & $3.742\times10^{-2}$ \\
 &  & p & $1.182\times10^{-4}$ & $1.087\times10^{-2}$ & $8.860\times10^{-3}$ & $1.224\times10^{-1}$ \\
 &  & T & $4.825\times10^{-4}$ & $2.197\times10^{-2}$ & $1.765\times10^{-2}$ & $4.162\times10^{-2}$ \\
 &  & phi & $1.153\times10^{-5}$ & $3.396\times10^{-3}$ & $2.163\times10^{-3}$ & $1.212\times10^{-2}$ \\
 &  & Avg. & $1.642\times10^{-4}$ & $1.133\times10^{-2}$ & $9.014\times10^{-3}$ & $4.868\times10^{-2}$ \\
\midrule
Case 3 & RA-PINN & u & $\mathbf{8.172\times10^{-9}}$ & $\mathbf{9.040\times10^{-5}}$ & $\mathbf{6.840\times10^{-5}}$ & $\mathbf{1.314\times10^{-4}}$ \\
 &  & v & $\mathbf{5.356\times10^{-9}}$ & $\mathbf{7.318\times10^{-5}}$ & $\mathbf{5.582\times10^{-5}}$ & $\mathbf{2.423\times10^{-2}}$ \\
 &  & p & $\mathbf{5.692\times10^{-9}}$ & $\mathbf{7.545\times10^{-5}}$ & $\mathbf{5.853\times10^{-5}}$ & $\mathbf{5.974\times10^{-4}}$ \\
 &  & T & $\mathbf{1.105\times10^{-8}}$ & $\mathbf{1.051\times10^{-4}}$ & $\mathbf{8.262\times10^{-5}}$ & $\mathbf{1.972\times10^{-4}}$ \\
 &  & phi & $5.327\times10^{-9}$ & $7.299\times10^{-5}$ & $6.653\times10^{-5}$ & $1.650\times10^{-4}$ \\
 &  & Avg. & $\mathbf{7.119\times10^{-9}}$ & $\mathbf{8.343\times10^{-5}}$ & $\mathbf{6.638\times10^{-5}}$ & $\mathbf{5.065\times10^{-3}}$ \\
\midrule
 & LSTM-PINN & u & $1.175\times10^{-8}$ & $1.084\times10^{-4}$ & $8.424\times10^{-5}$ & $1.575\times10^{-4}$ \\
 &  & v & $1.080\times10^{-8}$ & $1.039\times10^{-4}$ & $7.332\times10^{-5}$ & $3.442\times10^{-2}$ \\
 &  & p & $7.936\times10^{-9}$ & $8.908\times10^{-5}$ & $6.603\times10^{-5}$ & $7.054\times10^{-4}$ \\
 &  & T & $3.536\times10^{-8}$ & $1.880\times10^{-4}$ & $1.525\times10^{-4}$ & $3.528\times10^{-4}$ \\
 &  & phi & $\mathbf{4.072\times10^{-9}}$ & $\mathbf{6.381\times10^{-5}}$ & $\mathbf{4.991\times10^{-5}}$ & $\mathbf{1.442\times10^{-4}}$ \\
 &  & Avg. & $1.398\times10^{-8}$ & $1.107\times10^{-4}$ & $8.520\times10^{-5}$ & $7.155\times10^{-3}$ \\
\midrule
 & pLSTM-PINN & u & $3.674\times10^{-4}$ & $1.917\times10^{-2}$ & $1.566\times10^{-2}$ & $2.786\times10^{-2}$ \\
 &  & v & $1.531\times10^{-4}$ & $1.237\times10^{-2}$ & $9.865\times10^{-3}$ & $4.097\times10^{0}$ \\
 &  & p & $4.546\times10^{-5}$ & $6.743\times10^{-3}$ & $5.156\times10^{-3}$ & $5.339\times10^{-2}$ \\
 &  & T & $7.499\times10^{-5}$ & $8.659\times10^{-3}$ & $7.039\times10^{-3}$ & $1.625\times10^{-2}$ \\
 &  & phi & $2.186\times10^{-4}$ & $1.478\times10^{-2}$ & $1.289\times10^{-2}$ & $3.342\times10^{-2}$ \\
 &  & Avg. & $1.719\times10^{-4}$ & $1.235\times10^{-2}$ & $1.012\times10^{-2}$ & $8.456\times10^{-1}$ \\
\midrule
 & Pure-MLP & u & $5.577\times10^{-7}$ & $7.468\times10^{-4}$ & $5.806\times10^{-4}$ & $1.085\times10^{-3}$ \\
 &  & v & $1.835\times10^{-7}$ & $4.284\times10^{-4}$ & $3.373\times10^{-4}$ & $1.418\times10^{-1}$ \\
 &  & p & $3.348\times10^{-7}$ & $5.786\times10^{-4}$ & $4.378\times10^{-4}$ & $4.582\times10^{-3}$ \\
 &  & T & $3.020\times10^{-6}$ & $1.738\times10^{-3}$ & $1.183\times10^{-3}$ & $3.261\times10^{-3}$ \\
 &  & phi & $1.211\times10^{-7}$ & $3.480\times10^{-4}$ & $2.020\times10^{-4}$ & $7.866\times10^{-4}$ \\
 &  & Avg. & $8.434\times10^{-7}$ & $7.679\times10^{-4}$ & $5.481\times10^{-4}$ & $3.031\times10^{-2}$ \\
\midrule
Case 4 & RA-PINN & u & $9.196\times10^{-8}$ & $3.032\times10^{-4}$ & $2.173\times10^{-4}$ & $1.937\times10^{-3}$ \\
 &  & v & $1.329\times10^{-7}$ & $3.645\times10^{-4}$ & $2.774\times10^{-4}$ & $1.599\times10^{-3}$ \\
 &  & p & $6.623\times10^{-8}$ & $2.573\times10^{-4}$ & $1.826\times10^{-4}$ & $1.598\times10^{-3}$ \\
 &  & T & $1.419\times10^{-7}$ & $3.767\times10^{-4}$ & $2.977\times10^{-4}$ & $7.539\times10^{-4}$ \\
 &  & phi & $\mathbf{5.930\times10^{-8}}$ & $\mathbf{2.435\times10^{-4}}$ & $\mathbf{1.817\times10^{-4}}$ & $\mathbf{9.959\times10^{-4}}$ \\
 &  & Avg. & $\mathbf{9.845\times10^{-8}}$ & $\mathbf{3.091\times10^{-4}}$ & $\mathbf{2.313\times10^{-4}}$ & $\mathbf{1.377\times10^{-3}}$ \\
\midrule
 & LSTM-PINN & u & $\mathbf{7.200\times10^{-8}}$ & $\mathbf{2.683\times10^{-4}}$ & $\mathbf{2.023\times10^{-4}}$ & $\mathbf{1.714\times10^{-3}}$ \\
 &  & v & $\mathbf{1.160\times10^{-7}}$ & $\mathbf{3.406\times10^{-4}}$ & $\mathbf{2.614\times10^{-4}}$ & $\mathbf{1.494\times10^{-3}}$ \\
 &  & p & $\mathbf{4.899\times10^{-8}}$ & $\mathbf{2.213\times10^{-4}}$ & $\mathbf{1.697\times10^{-4}}$ & $\mathbf{1.374\times10^{-3}}$ \\
 &  & T & $\mathbf{7.197\times10^{-8}}$ & $\mathbf{2.683\times10^{-4}}$ & $\mathbf{2.052\times10^{-4}}$ & $\mathbf{5.369\times10^{-4}}$ \\
 &  & phi & $2.706\times10^{-7}$ & $5.202\times10^{-4}$ & $3.217\times10^{-4}$ & $2.127\times10^{-3}$ \\
 &  & Avg. & $1.159\times10^{-7}$ & $3.238\times10^{-4}$ & $2.321\times10^{-4}$ & $1.449\times10^{-3}$ \\
\midrule
 & pLSTM-PINN & u & $4.012\times10^{-5}$ & $6.334\times10^{-3}$ & $4.826\times10^{-3}$ & $4.047\times10^{-2}$ \\
 &  & v & $2.605\times10^{-5}$ & $5.104\times10^{-3}$ & $3.864\times10^{-3}$ & $2.239\times10^{-2}$ \\
 &  & p & $6.524\times10^{-5}$ & $8.077\times10^{-3}$ & $6.195\times10^{-3}$ & $5.015\times10^{-2}$ \\
 &  & T & $1.181\times10^{-4}$ & $1.087\times10^{-2}$ & $8.368\times10^{-3}$ & $2.175\times10^{-2}$ \\
 &  & phi & $2.153\times10^{-4}$ & $1.467\times10^{-2}$ & $1.168\times10^{-2}$ & $6.000\times10^{-2}$ \\
 &  & Avg. & $9.296\times10^{-5}$ & $9.011\times10^{-3}$ & $6.986\times10^{-3}$ & $3.895\times10^{-2}$ \\
\midrule
 & Pure-MLP & u & $6.785\times10^{-6}$ & $2.605\times10^{-3}$ & $1.985\times10^{-3}$ & $1.664\times10^{-2}$ \\
 &  & v & $5.796\times10^{-6}$ & $2.408\times10^{-3}$ & $1.896\times10^{-3}$ & $1.056\times10^{-2}$ \\
 &  & p & $2.579\times10^{-6}$ & $1.606\times10^{-3}$ & $1.241\times10^{-3}$ & $9.972\times10^{-3}$ \\
 &  & T & $8.578\times10^{-6}$ & $2.929\times10^{-3}$ & $2.316\times10^{-3}$ & $5.862\times10^{-3}$ \\
 &  & phi & $6.000\times10^{-6}$ & $2.450\times10^{-3}$ & $1.538\times10^{-3}$ & $1.002\times10^{-2}$ \\
 &  & Avg. & $5.948\times10^{-6}$ & $2.399\times10^{-3}$ & $1.795\times10^{-3}$ & $1.061\times10^{-2}$ \\
\bottomrule
\end{longtable}
\endgroup

\noindent Table~\ref{tab:error_all} now lists the field-wise errors for $p$, $u$, $v$, $\phi$, and $T$ together with the corresponding Avg. rows used for the case-level comparison discussed below.
\FloatBarrier

\begin{table}[t]
\centering
\caption{Training time in hours converted from the logged time-cost files.}
\label{tab:time_all}
\small
\setlength{\tabcolsep}{6pt}
\begin{tabular}{cccccc}
\toprule
Case & RA-PINN & LSTM-PINN & pLSTM-PINN & Pure-MLP & Fastest \\
\midrule
Case 1 & 24.01 & 14.29 & 1.09 & 1.40 & pLSTM-PINN \\
Case 2 & 24.67 & 14.76 & 4.19 & 3.17 & Pure-MLP \\
Case 3 & 39.81 & 11.83 & 4.28 & 4.45 & pLSTM-PINN \\
Case 4 & 38.35 & 18.35 & 9.30 & 14.76 & pLSTM-PINN \\
\bottomrule
\end{tabular}
\end{table}
\FloatBarrier

\section{Case 1: Constant-Coefficient Coupled Benchmark}
Case~1 uses the baseline constant-coefficient electrothermal coupled system. We keep the unified interior residual $\mathcal{N}(\bU)$ defined in \Eq{eq:operator} and take the transport coefficients and electrothermal coupling parameters as spatially uniform constants. The case residual therefore consists of the continuity residual in \Eq{eq:cont}, the two momentum residuals in \Eq{eq:mom}, the electric-potential residual in \Eq{eq:phi_eq}, and the temperature residual in \Eq{eq:T_eq}, together with full Dirichlet boundary residuals for $u$, $v$, $p$, $\phi$, and $T$. This construction makes Case~1 the cleanest benchmark in the group, because the coupling remains fully active while the coefficient law does not introduce additional nonlinearity or interface jumps.

\begin{figure}[t]
    \centering
    \includegraphics[width=0.97\textwidth,height=0.73\textheight,keepaspectratio]{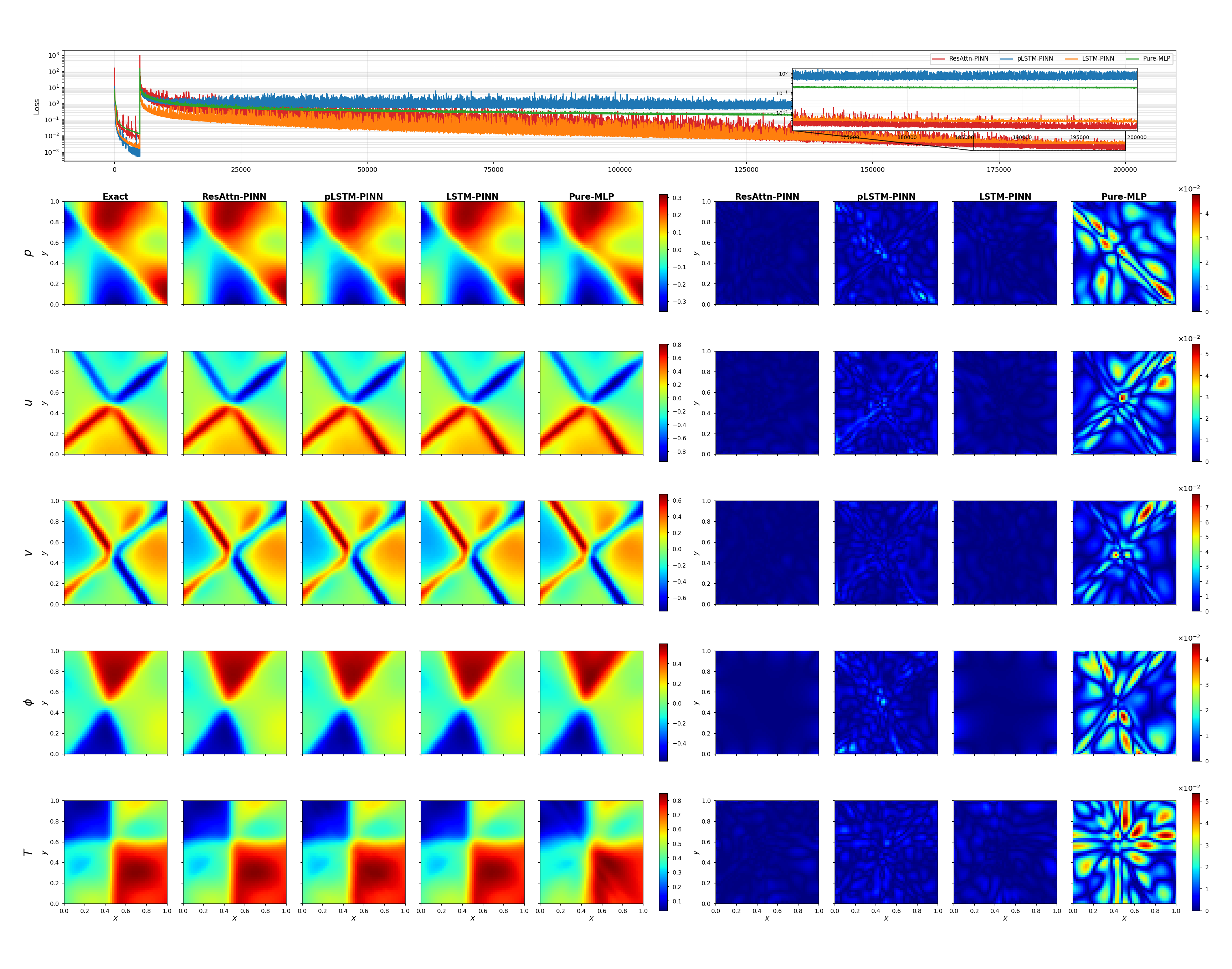}
    \caption{Uploaded benchmark comparison for Case~1. The top strip reports the loss-history comparison, the five rows correspond to $p$, $u$, $v$, $\phi$, and $T$, the left block reports the benchmark and predicted fields, and the right block reports the absolute-error maps of the four learned solvers.}
    \label{fig:case1}
\end{figure}
\FloatBarrier

\Fig{fig:case1} shows that RA-PINN reconstructs all five fields with the cleanest global geometry and the weakest background contamination. The pressure map preserves the smooth large-scale variation and the internal transition bands with very limited distortion. The velocity and potential channels also remain sharply organized around the steep diagonal structures, whereas the competing models either blur these layers or introduce visible oscillatory artifacts. Pure-MLP produces the strongest spatial noise over all five outputs, while pLSTM-PINN loses local sharpness more clearly than LSTM-PINN. The visual comparison therefore indicates that residual attention improves multi-field coordination even in the simplest coefficient regime.

The quantitative comparison in \Tab{tab:error_all} confirms the same ordering. In Case~1, RA-PINN attains the minimum values for MSE, RMSE, MAE, and relative $L_2$ error in all five physical fields and also in the Avg. row. Relative to LSTM-PINN, the averaged MSE decreases from $2.901\times10^{-6}$ to $9.083\times10^{-7}$, and the averaged relative $L_2$ error decreases from $5.695\times10^{-3}$ to $3.235\times10^{-3}$. The gap becomes much larger relative to pLSTM-PINN and Pure-MLP, whose field-wise and averaged errors remain substantially higher. \Tab{tab:time_all} also shows that this accuracy gain does not come with a time advantage, because RA-PINN requires 24.01~h whereas pLSTM-PINN and Pure-MLP finish in about one hour. For this constant-coefficient case, the tabulated evidence therefore supports a clear accuracy improvement from RA-PINN, but it also shows a substantial increase in training cost.

\section{Case 2: Pressure-Gauge Benchmark}
Case~2 keeps the same coupled five-field operator, but the pressure treatment changes. We still minimize the continuity, momentum, electric-potential, and temperature residuals in \Eq{eq:operator}, while the pressure constraint replaces direct pressure anchoring by the gauge residual associated with \Eq{eq:gauge}. In practice, the residual construction now combines exact Dirichlet boundary residuals for $u$, $v$, $\phi$, and $T$ with a zero-mean pressure condition over the whole domain. This modification removes direct pointwise pressure prescription and forces the solver to recover the pressure level from the coupled operator itself.

\begin{figure}[t]
    \centering
    \includegraphics[width=0.97\textwidth,height=0.73\textheight,keepaspectratio]{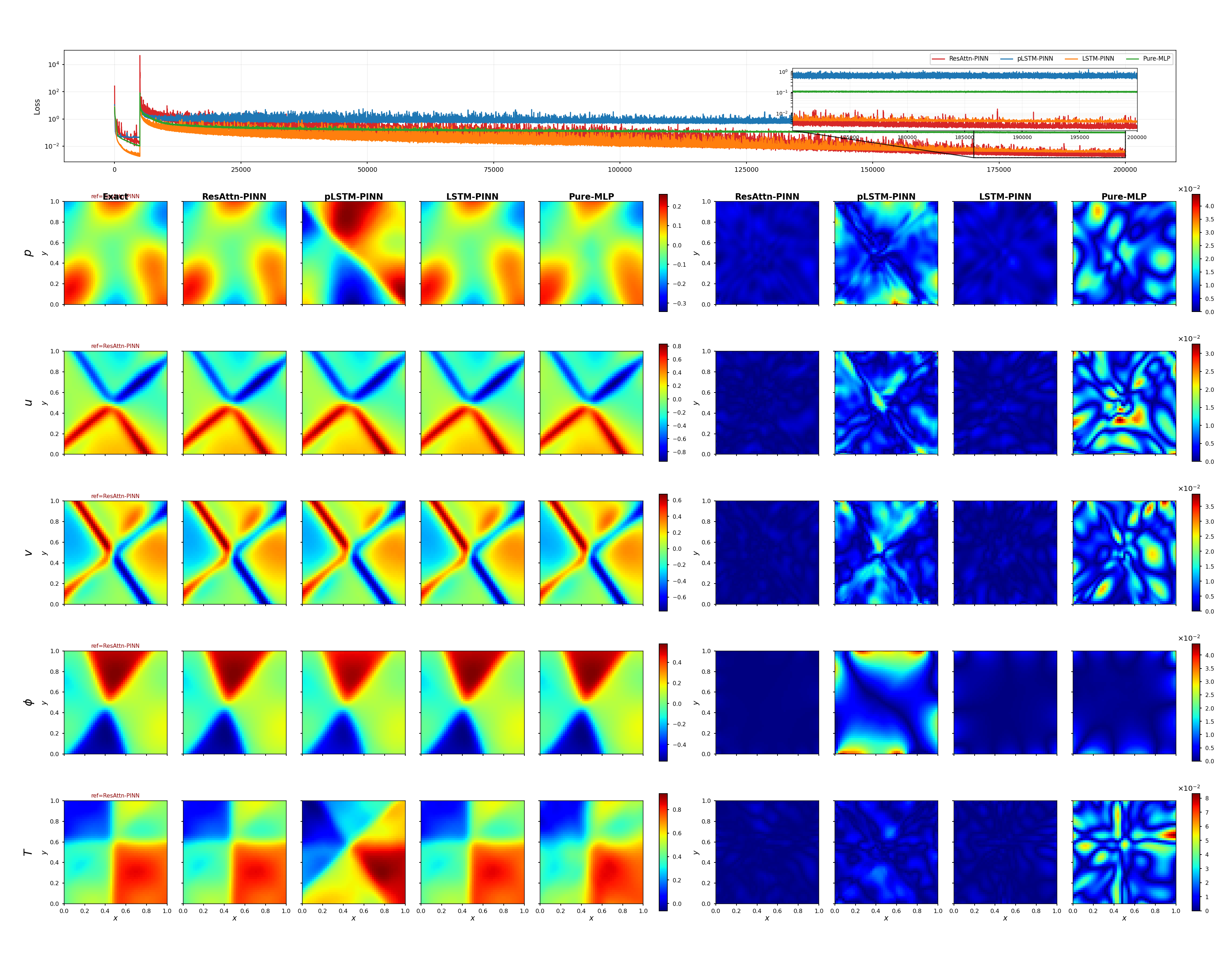}
    \caption{Uploaded benchmark comparison for Case~2. This benchmark replaces direct pressure anchoring with a zero-mean gauge constraint, so the figure tests how each network reconstructs the five coupled fields under indirect pressure identifiability.}
    \label{fig:case2}
\end{figure}
\FloatBarrier

\Fig{fig:case2} shows that the pressure-gauge treatment increases the reconstruction difficulty for all models. The benchmark canvas explicitly notes an exact mismatch warning, so we mainly use the figure for structural comparison rather than absolute reading of every color level. Even under this stricter setting, RA-PINN still preserves the most coherent field organization, especially in the pressure and temperature rows, where the competing models show broader deformation and more persistent structured artifacts. LSTM-PINN remains the second-best solver, pLSTM-PINN exhibits strong low-frequency contamination, and Pure-MLP accumulates the largest global error pattern.

\Tab{tab:error_all} again gives the same ranking. In Case~2, RA-PINN retains the minimum values in all five physical fields and also in the Avg. row. The reduction relative to LSTM-PINN remains moderate but consistent in every averaged column, for example RMSE decreases from $1.902\times10^{-3}$ to $1.235\times10^{-3}$ and relative $L_2$ error decreases from $1.038\times10^{-2}$ to $7.660\times10^{-3}$. The advantage becomes much larger relative to pLSTM-PINN and Pure-MLP, both of which show visibly larger field-wise and averaged errors under the gauge constraint. \Tab{tab:time_all} indicates that RA-PINN still needs the longest training time in this case at 24.67~h, whereas Pure-MLP finishes in 3.17~h and pLSTM-PINN in 4.19~h. The table evidence therefore suggests that RA-PINN improves pressure-gauge reconstruction accuracy in a stable way, but it does so with a clear computational-time penalty.

\section{Case 3: Temperature-Dependent Transport Benchmark}
Case~3 activates the variable-coefficient residual defined by the temperature-dependent constitutive law in \Eq{eq:tempdep}. We still use the unified operator $\mathcal{N}(\bU)$, but the momentum and temperature residuals now depend on transport coefficients that change with the predicted temperature. The residual construction therefore becomes more strongly nonlinear than in the first two cases, because temperature error feeds back into the viscous and thermal transport operators themselves. We keep the same coupled field set $\{u,v,p,\phi,T\}$ and the same boundary-consistency philosophy, so the only new ingredient comes from the coefficient-response mechanism.

\begin{figure}[t]
    \centering
    \includegraphics[width=0.97\textwidth,height=0.73\textheight,keepaspectratio]{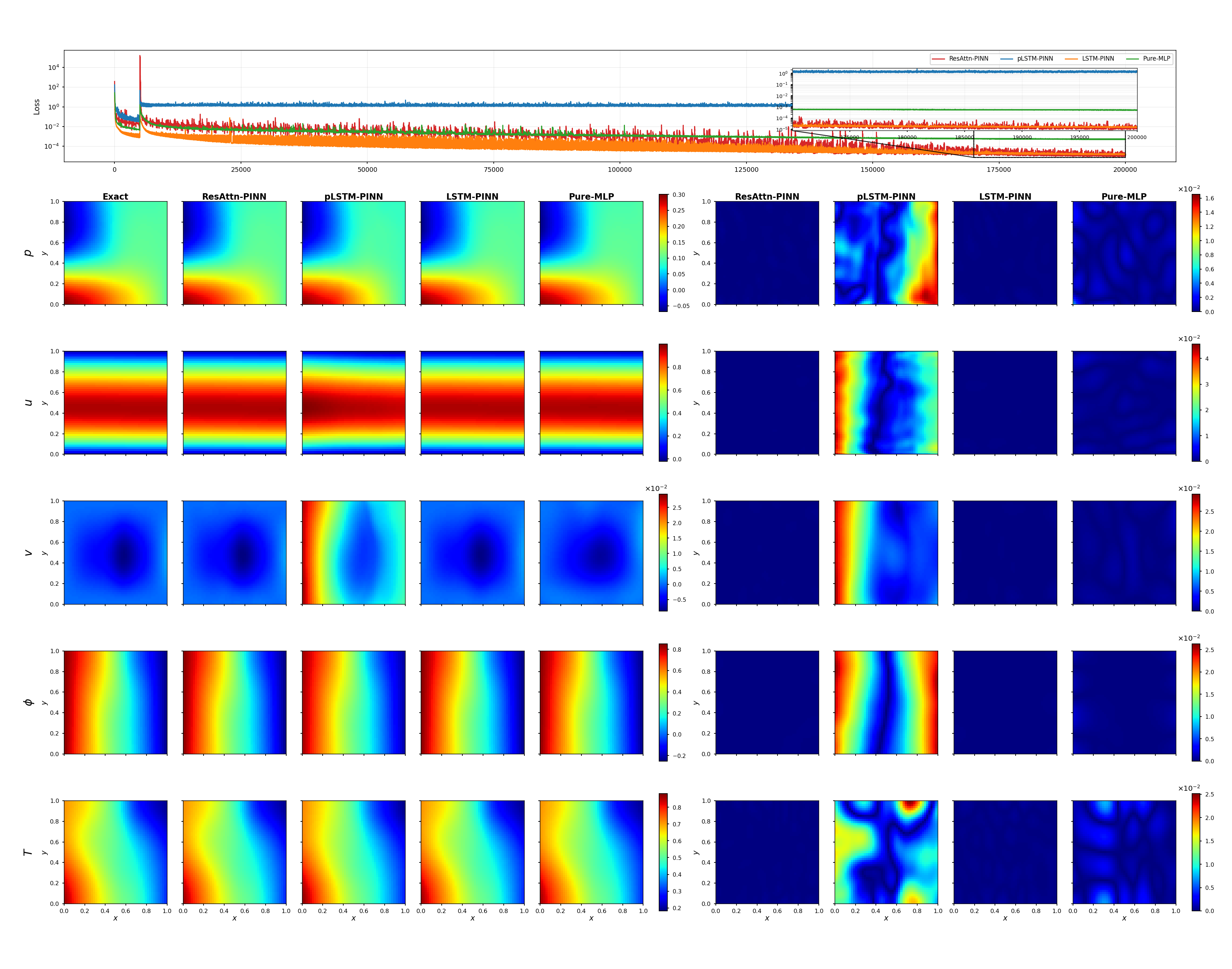}
    \caption{Uploaded benchmark comparison for Case~3. This benchmark activates temperature-dependent transport coefficients, so the figure directly tests how well each network resolves variable-coefficient coupling across the five physical fields.}
    \label{fig:case3}
\end{figure}
\FloatBarrier

\Fig{fig:case3} shows a particularly clear separation between the four solvers. RA-PINN almost overlaps with the benchmark field pattern across all five rows and keeps the smooth field geometry intact without introducing visible strip-like artifacts. LSTM-PINN also performs well, but it leaves slightly larger residual traces in several channels. Pure-MLP reproduces the broad tendency of the solution, yet it loses fine-scale accuracy more clearly. pLSTM-PINN deteriorates most strongly here and generates large structured distortions, which indicates that variable-coefficient transport changes the preferred inductive bias of the solver in a way that the recurrent-parallel backbone cannot absorb effectively.

The metrics in \Tab{tab:error_all} support the same conclusion, but the field-wise breakdown adds one useful nuance. In Case~3, RA-PINN gives the minimum values for $u$, $v$, $p$, and $T$ across all four metrics and also attains the best Avg. row, whereas LSTM-PINN is slightly better in the $\phi$ field. Even with that local exception, the overall ranking remains unchanged: the averaged relative $L_2$ error stays at $5.065\times10^{-3}$ for RA-PINN, compared with $7.155\times10^{-3}$ for LSTM-PINN, $3.031\times10^{-2}$ for Pure-MLP, and $8.456\times10^{-1}$ for pLSTM-PINN. This means that the variable-coefficient case amplifies the difference between solvers much more strongly than the previous two benchmarks. \Tab{tab:time_all} shows the cost side of the same result: RA-PINN requires 39.81~h, which is the longest runtime among all reported entries in the paper. The table comparison therefore indicates that RA-PINN remains the most reliable solver for temperature-dependent transport, but the added robustness comes with the highest computational expense in the full benchmark set.

\section{Case 4: Oblique-Interface Benchmark}
Case~4 introduces the interface residual associated with the oblique geometry in \Eq{eq:interface} and the jump-condition operator in \Eq{eq:jump}. We keep the same interior residual vector $\mathcal{N}(\bU)$ inside each subdomain, assign different material coefficients on the two sides of the slanted interface, and add interface residual terms that enforce continuity of $u$, $v$, $p$, $\phi$, and $T$ together with the prescribed flux-jump consistency. The residual construction is therefore the richest among the four benchmarks, because the solver must coordinate outer-boundary constraints, interior PDE balance, and interface transmission conditions at the same time.

\begin{figure}[t]
    \centering
    \includegraphics[width=0.97\textwidth,height=0.73\textheight,keepaspectratio]{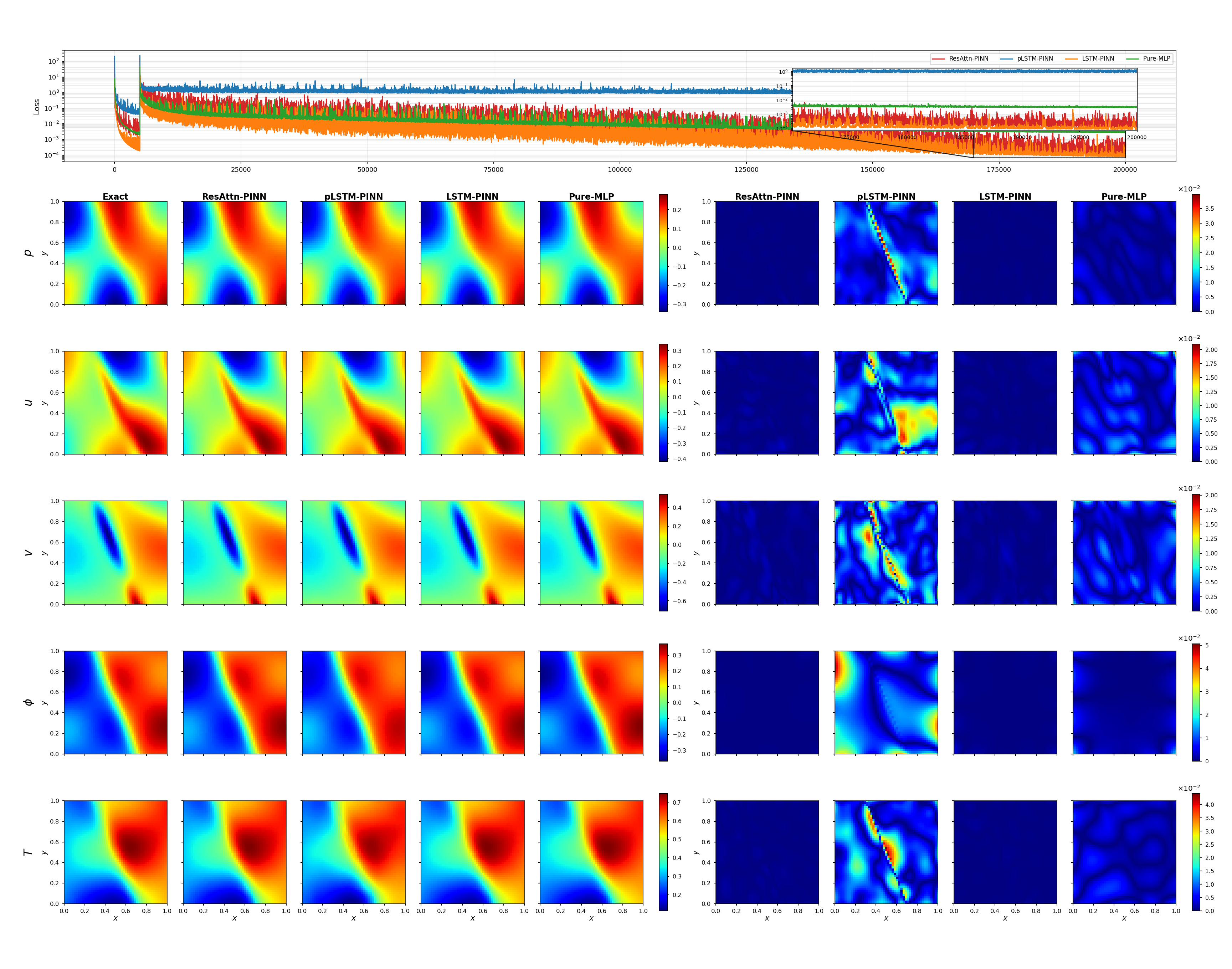}
    \caption{Uploaded benchmark comparison for Case~4. The benchmark contains an oblique material interface, so the figure highlights interface-sensitive reconstruction quality by placing the learned absolute-error maps next to the predicted multiphysics fields.}
    \label{fig:case4}
\end{figure}
\FloatBarrier

\Fig{fig:case4} shows that RA-PINN resolves the slanted interface most cleanly and keeps the surrounding background field smooth at the same time. LSTM-PINN approaches RA-PINN closely, but it still leaves slightly thicker residual bands around the interface. pLSTM-PINN generates strong interface-aligned oscillations, and Pure-MLP produces broader nonphysical patterns that spread into the surrounding domain. Because the interface does not align with the Cartesian axes, this benchmark strongly rewards architectures that can preserve localized geometric information without oversmoothing it.

\Tab{tab:error_all} shows that Case~4 is the closest comparison in the whole benchmark set. The field-wise rows indicate that LSTM-PINN is slightly better for $u$, $v$, $p$, and $T$, whereas RA-PINN is clearly better for $\phi$ and still retains the best Avg. row across MSE, RMSE, MAE, and relative $L_2$. In particular, the averaged MAE changes only slightly from $2.321\times10^{-4}$ for LSTM-PINN to $2.313\times10^{-4}$ for RA-PINN, and the averaged relative $L_2$ error decreases from $1.449\times10^{-3}$ to $1.377\times10^{-3}$. The table therefore shows that the interface-dominated setting produces a much narrower margin than the previous cases. \Tab{tab:time_all} also confirms that RA-PINN is not the most efficient option here, because it requires 38.35~h compared with 18.35~h for LSTM-PINN and 9.30~h for pLSTM-PINN. For the oblique-interface case, the tabulated evidence supports a modest overall accuracy gain from RA-PINN at a substantially higher training cost.

\section*{Supplementary Material}
Supplementary material accompanies this article. It now contains the extended technical support items omitted from the main text: notation and case activation in \SMSec{I}, residual-attention solver decomposition in \SMSec{II}, unified residual closure in \SMSec{III}, detailed loss channels and case-wise reduction in \SMSec{IV}, adaptive collocation and interface sampling in \SMSec{V}, benchmark metric definitions in \SMSec{VI}, supporting literature inventory in \SMSec{VII}, and the theory-to-implementation correspondence in \SMSec{VIII}.

\section{Conclusion}
This study demonstrates that the RA-PINN framework provides a robust and high-precision computational approach for solving steady-state electrothermal coupled problems within a unified partial differential equation formulation. Through extensive validation across four distinct benchmarks, the proposed architecture consistently achieves the highest overall average accuracy compared to established neural solvers such as Pure-MLP, LSTM-PINN, and pLSTM-PINN. The performance advantage of RA-PINN becomes particularly pronounced in scenarios involving strong nonlinear coupling and temperature-dependent variable coefficients, where conventional backbones often struggle to maintain structural fidelity. While the interface-dominated case results in a narrower margin of improvement over specialized recurrent models, the residual-attention mechanism effectively preserves the sharp transitions and localized features critical for reliable energy system modeling. These results confirm that integrating attention-guided channel modulation with adaptive sampling significantly enhances the solver's ability to resolve the complex multiphysics interactions found in modern thermal management applications. Although the current implementation requires a higher computational investment during the training phase, the gains in solution reliability establish a strong foundation for future research aimed at optimizing training efficiency without compromising numerical precision. Consequently, the RA-PINN framework serves as a promising tool for the high-fidelity simulation and optimization of next-generation electrothermal energy devices.

\printcredits

\section*{Declaration of competing interest}
The authors declared that they have no conflicts of interest to this work. 
\section*{Acknowledgment}
This work is supported by the developing Project of Science and Technology of Jilin Province (20250102032JC). 

\section*{Data availability}
All the code for this article is available open access at a Github repository available at https://github.com/Uderwood-TZ/RA-PINN-Application-of-RA-PINN-in-Solving-Steady-State-Electrothermal-Coupled-Multiphysics-Problems.

\bibliographystyle{unsrtnat}
\bibliography{references}

\end{document}